\pdfoutput=1

\documentclass[11pt]{article}

\usepackage[final]{acl}

\usepackage{times}
\usepackage{latexsym}

\usepackage{xspace}
\usepackage{subcaption}
\usepackage{multirow}
\usepackage{CJKutf8}

\usepackage[T1]{fontenc}

\usepackage[utf8]{inputenc}

\usepackage{microtype}

\usepackage{inconsolata}

\usepackage{graphicx}
\usepackage{colortbl}
\usepackage{xcolor}
\usepackage{booktabs}
\usepackage{graphicx}

\usepackage{enumitem}

\newcommand{\appname}{\textsc{TranslationCorrect}\xspace}

\title{\appname: A Unified Framework for Machine Translation Post-Editing with Predictive Error Assistance}

\author{
 Syed Mekael Wasti\textsuperscript{*, \textdagger} \
 Shou-Yi Hung\textsuperscript{*, \textdaggerdbl} \
 Christopher Collins\textsuperscript{\textdagger} \
 En-Shiun Annie Lee\textsuperscript{\textdagger, \textdaggerdbl} \\
\textsuperscript{\textdagger}Ontario Tech University \
\textsuperscript{\textdaggerdbl}University of Toronto \\
\texttt{syedmekael.wasti@ontariotechu.net}, \texttt{sy.hung@mail.utoronto.ca}
}

\begin{document}
\maketitle

\def\thefootnote{*}\footnotetext{Equal contribution, corresponding author}

\def\thefootnote{\arabic{footnote}}

\begin{abstract}

Machine translation (MT) post-editing and research data collection often rely on inefficient, disconnected workflows. We introduce \appname, an integrated framework designed to streamline these tasks. \appname combines MT generation using models like NLLB, automated error prediction using models like XCOMET or LLM APIs (providing detailed reasoning), and an intuitive post-editing interface within a single environment. Built with human-computer interaction (HCI) principles in mind to minimize cognitive load, \appname makes it easier for annotators to perform annotations, as confirmed by a user study using NASA Task Load Indices. For translators, it enables them to correct errors and batch translate efficiently. For researchers, \appname exports high-quality span-based annotations in the Error Span Annotation (ESA) format, using an error taxonomy inspired by Multidimensional Quality Metrics (MQM). These outputs are compatible with state-of-the-art error detection models and suitable for training MT or post-editing systems. Our user study confirms that \appname significantly improves translation efficiency and user satisfaction over traditional annotation methods.

\end{abstract}

\section{Introduction}

Machine translation (MT) has seen significant advancements with the development of powerful translation models like Meta's No Language Left Behind \citep[NLLB]{nllbteam2022languageleftbehindscaling} and evaluation tools such as XCOMET \cite{guerreiro-etal-2024-xcomet}. However, the current translation and data collection workflows for MT model training remain inefficient. Traditional translation procedures often require human annotators to rely on manual, time-consuming processes involving tools like CSV files or Excel sheets \cite{federmann-2018-appraise}. Typically, a translator must first generate machine translations using an external model, then manually collect and transfer the output into another format for review. Any subsequent error correction must also be performed manually, resulting in an inefficient and error-prone process. 

Similar challenges also exist in the data collection process for MT research. Datasets used for training MT systems are often complex to collect, as they have to undergo the tedious manual process mentioned earlier. However, to improve the efficiency of the annotation process, annotation tools like Appraise \cite{federmann-2018-appraise} have been developed to facilitate the whole process, making MT training data collection easier and standardizing the data collection procedure. However, Appraise remains a platform dedicated to experienced annotators and linguists, enabling them to annotate data for future research and model training, which limits its usage to a specific group of users. 

\begin{figure*}[t!]
    \centering
    \includegraphics[width=1\linewidth]{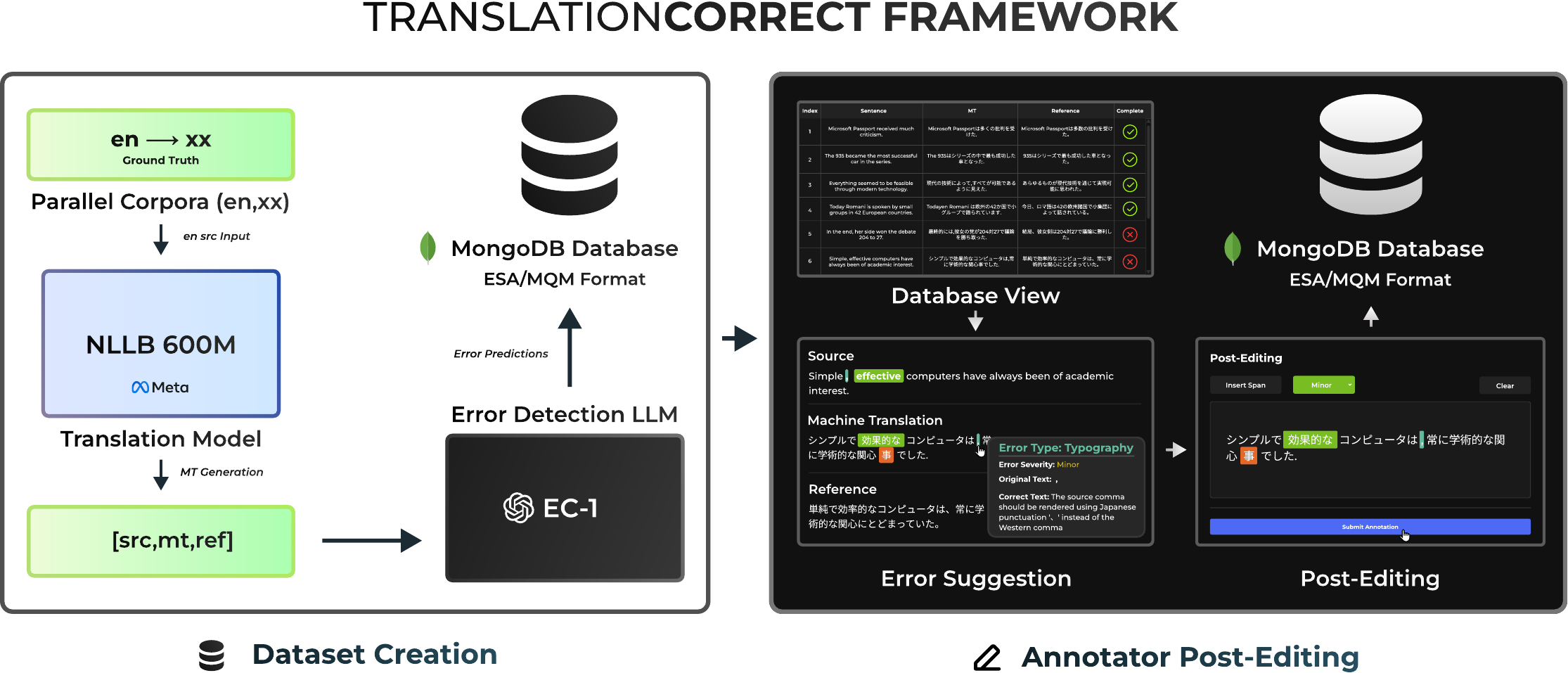}
    \caption{Overview of the \appname framework. The workflow begins with an annotator fetching data from a previously populated database we create for en$\rightarrow$xx language sets. We process our collections using the EC-1 error detection model and analyze the MT output to identify potential errors. The user sees these selected sentences and the relevant predicted errors within the post-edit section, where they can correct the translation based on these suggestions before submitting the final annotated sentence.}
    \label{fig:flowchart}
\end{figure*}

To address these limitations, we introduce \appname, a framework designed to streamline both translation workflows and MT data collection. For translators, \appname offers a solution that automatically generates initial translations using a translation model, such as NLLB and identifies potential translation errors using XCOMET or an LLM of choice to provide more insights into the translation errors, enabling efficient post-editing of translations within the same environment. This approach eliminates the need for manual data handling through external tools, improving both translation quality and efficiency. For researchers in the MT community, \appname also serves as a robust data collection tool, automatically formatting outputs in alignment with state-of-the-art MT dataset standards, supporting outputs that contain Multidimensional Quality Metrics \citep[MQM]{burchardt-2013-multidimensional-mqm} and Error Span Annotation \citep[ESA]{kocmi-etal-2024-error-esa} information alongside each translation source and target pair. This feature enables annotators to generate high-quality datasets that can be used directly for training error correction models like XCOMET or fine-tuning translation systems like NLLB. 

Furthermore, our framework is designed with human-computer interaction (HCI) principles in mind, prioritizing ease of use and flexibility for annotators. The user interface is designed to minimize cognitive load and reduce the difficulty typically associated with traditional annotation workflows, such as those relying on manual data processing \cite{hci-design-principles, hci-principles-of-usability} through Microsoft Excel. By integrating MT generation, error prediction, and correction within a single environment, our framework enables translators to focus on the translation task itself, rather than having to work with multiple tools simultaneously. Evaluation results from a user study indicate that our framework significantly outperforms traditional annotation methods, resulting in a considerably lower perceived workload and increased efficiency compared to conventional annotation methods.

Our contributions are summarized as follows:

\begin{itemize}
    \item \appname offers an integrated environment that automatically generates initial translations using translation models, predicts potential errors using error detection models or an LLM of choice, and enables efficient corrections.
    \item The framework supports output formats aligned with state-of-the-art MT dataset standards, including MQM and ESA, enabling researchers and annotators to generate high-quality datasets for training and fine-tuning MT and translation error detection models.
    \item Designed with HCI principles in mind, \appname prioritizes ease of use and flexibility, reducing cognitive load for annotators. 
\end{itemize}

Our repository is MIT Licensed and is publicly available on GitHub\footnote{\href{https://github.com/MekaelWasti/TranslationCorrect}{https://github.com/MekaelWasti/TranslationCorrect}}. Our deployed demo is available on a website\footnote{\href{https://translation-correct-annotation-git-27a7e8-mekaelwastis-projects.vercel.app/}{https://translation-correct-annotation-git-27a7e8-mekaelwastis-projects.vercel.app/}}. A short demo of our framework is available on YouTube\footnote{\href{https://youtu.be/j2sp13qyeQM}{https://youtu.be/j2sp13qyeQM}}.

\section{\appname}

An overview of \appname 's workflow is illustrated in Figure~\ref{fig:flowchart}, outlining the user flow from target language dataset selection to automatic error detection, user post-editing, and data export.

\begin{figure}
    \centering
    \includegraphics[width=\linewidth]{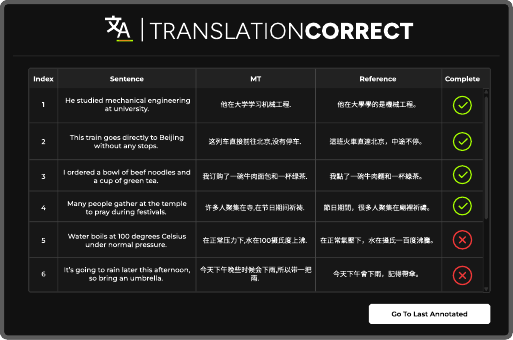}
    \caption{Annotators can use the database view to easily view their target language dataset, select their desired sentences, and monitor completion status}
    \label{fig:database-view}
\end{figure}

\subsection{Database View \& MT Generation}



\begin{figure}[b!]
    \centering
    \includegraphics[width=\linewidth]{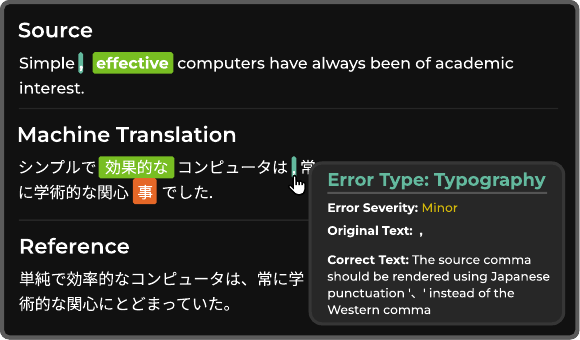}
    \caption{Predicted errors annotated by our error detection model are highlighted in both the source text and the machine translation output, with a detailed description of the error identified and its source-to-MT mapping.}
    \label{fig:error-highlight}
\end{figure}

When annotators first enter the \appname interface, they are presented with a database view component to select from the data entries that require annotation. Annotators can load sentences from this view, containing multiple source and MT pairs. The dataset is stored in MongoDB, containing the precomputed pairs of source and machine-translated sentences. For most of our datasets, we have used a 600M NLLB model\footnote{\href{https://huggingface.co/facebook/nllb-200-distilled-600M}{https://huggingface.co/facebook/nllb-200-distilled-600M}} to create machine translations; however, as we add increasingly lower-resource languages, we can switch to other models that support them. The source text and translated output are displayed side by side, as shown in Figure~\ref{fig:error-highlight}, allowing annotators to compare and assess the machine translation quality conveniently. Furthermore, the annotated data is saved to the same database, permitting easy access.


\subsection{Error Detection}

Following the MT generation, \appname integrates an error detection model of the user's choice to identify potential errors in the translated output automatically. For our demo, we offer two methods, with the first being XCOMET-XL\footnote{\href{https://huggingface.co/Unbabel/XCOMET-XL}{https://huggingface.co/Unbabel/XCOMET-XL}}, the 3.5B parameter variant of XCOMET, which will be used as a baseline error detection model, and the other being a custom GPT-4o assistant named EC-1. 




\subsubsection{Custom GPT-4o EC-1 Assistant}

We offer the option to apply a custom GPT-4o assistant, EC-1, to help users identify potential errors with an in-depth explanation, as shown in Figure~\ref{fig:error-highlight}. EC-1 is model-agnostic; other LLMs capable of structured JSON error-span output could be used in place of GPT-4o. However, 4o is cost-effective, fast and performs error detection with consistent accuracy compared to the tested OpenAI models. We leverage this model as an error detection model, using prompt engineering techniques to ensure that the response provided by our custom-crafted EC-1 assistant aligns with our standardized error ruleset, as outlined in Appendix~\ref{sec:appendix_error_definition}. As shown in Figure~\ref{fig:error-highlight}, translation errors are highlighted in different colors, present in both the source sentence and the MT output, allowing users to identify potential errors with minimal effort. Furthermore, a detailed explanation of the error is displayed when the user hovers their cursor above the highlighted text. Our human study shows that this provides a more in-depth analysis than using only the XCOMET model. 


The EC-1 assistant's response is obtained from an API endpoint, allowing it to be used in minimal client-side and limited computing environments without requiring additional computational resources to run local models; however, API calls to GPT-4o may incur large cloud usage costs depending on usage and the size of input datasets. Implementation details of our custom EC-1 model can be found in Appendix~\ref{sec:appendix_gpt_4o}. 


\subsection{User Post-Editing}

\begin{figure}
    \centering
    \includegraphics[width=\linewidth]{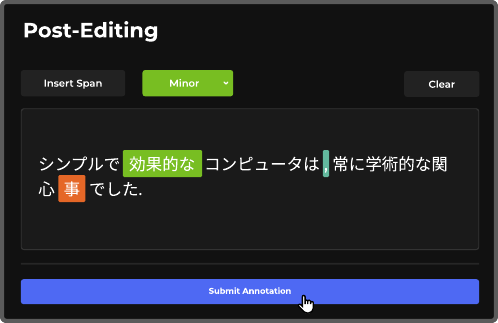}
    \caption{The Post-Edit component allows users to make detailed, fine-grained error edits on top of the potential error spans generated by our error detection model.}
    \label{fig:edit-context}
\end{figure}

Once errors are identified, users can correct translation errors directly within the system, as shown in Figure~\ref{fig:edit-context}. If the suggested errors do not match the user's expectations, they are allowed to make fine-grained edits to modify the detected errors and the final translated sentence. 

Users are also allowed to insert custom new error spans that the error detection model did not previously highlight.

\subsection{Data Export}

Once the post-editing process is completed, the user can export the final translation and annotations into a structured dataset. This feature allows one-click data export in a format compatible with MQM and ESA standards. 

The exported data can be downloaded from the interface in multiple formats, including CSV and JSON. It contains information on the source text, MT output, corrected text, error spans, error categories, and error severities. 

In addition to annotators manually downloading the data, the server manager can also fetch the annotated data from the MongoDB database connected to the server, allowing for easier management and exportation of the annotated data. 



\subsection{HCI Considerations}


To design an interface that reduces cognitive load, multiple HCI principles must work in tandem. \appname's interface is simple and clutter-free. This reduces the likelihood of annotators becoming overwhelmed or fatigued by unnecessary content on the screen. We ensured a strict workflow to minimize noise between annotators' submissions. 

A dark theme was chosen for the application to reduce visual fatigue from bright colors during long sessions. This was well received and praised by participants in our study. Additionally, vibrant and unique colors were chosen to represent different error categories, allowing users to quickly associate colors with categories, which is especially helpful when viewing error predictions. The interface also provides quick action shortcuts that appear near the annotator's cursor for crucial operations, such as inserting and deleting spans. The local pop-up reduces the distance required for mouse movement and speeds up the annotation process.

A comprehensive user study, further elaborated in the following section, confirms that users find the framework more effective, enjoyable, and efficient than traditional spreadsheet-based annotation workflows.

\begin{table*}[t]
\renewcommand{\arraystretch}{1.4}  

\centering
\resizebox{\linewidth}{!}{
\begin{tabular}{lcccccc}
\toprule
\textbf{Method} & \textbf{Mental ($\downarrow$)} & \textbf{Physical ($\downarrow$)} & \textbf{Temporal ($\downarrow$)} & \textbf{Performance ($\uparrow$)} & \textbf{Effort ($\downarrow$)} & \textbf{Frustration ($\downarrow$)} \\
\midrule
Excel & 4.10 ± 2.51 & 3.40 ± 2.88 & 2.70 ± 2.26 & 7.80 ± 1.55 & 4.10 ± 2.38 & 3.50 ± 2.92 \\
No Suggestions & 4.17 ± 2.52 & 2.42 ± 2.57 & 3.58 ± 2.02 & 8.58 ± 1.16 & 3.42 ± 1.16 & 1.83 ± 2.41 \\
XCOMET & 2.92 ± 1.56 & \textbf{1.58 ± 1.51} & 2.50 ± 1.68 & \textbf{8.67 ± 1.07} & \textbf{2.67 ± 1.07} & 1.92 ± 2.31 \\
EC-1& \textbf{2.67 ± 1.87} & \textbf{1.58 ± 1.08} & \textbf{2.17 ± 1.59} & 8.50 ± 1.00 & 3.08 ± 1.00 & \textbf{1.75 ± 2.26} \\
\bottomrule
\end{tabular}
}
\caption{Comparison of NASA TLX dimensions across annotation methods, with Excel annotations done following instructions outlined in Appendix~\ref{sec:appendix_excel}, and the different error detection settings used within \appname. Lower is better ($\downarrow$) for all metrics except Performance ($\uparrow$). Bold indicates the best score for each metric. }
\label{tab:nasa_tlx}
\end{table*}


\section{Results and Evaluation}






To evaluate the efficacy of \appname's interface design and the cognitive workload compared to manual annotation methods, a user study was conducted. The \textbf{NASA Task Load Index \citep[TLX]{nasa-taskload-index}} was used to measure workload across six dimensions: \textbf{Mental Demand}, \textbf{Physical Demand}, \textbf{Temporal Demand}, \textbf{Performance}, \textbf{Effort}, and \textbf{Frustration}. Overall, the results indicate that using \appname, particularly with our EC-1 error detection model, resulted in significantly lower perceived workload compared to the traditional Excel-based annotation method.

The participant pool comprised 12 annotators across 6 languages (Mandarin, Cantonese, Bengali, French, Japanese, Tamil). All annotators were native speakers of the respective non-English language and participated voluntarily. Details of the data collection process on the user study can be found in Appendix~\ref{sec:appendix_user_study_details}. The study was conducted under the following conditions:

\paragraph{User Study Conditions}

Each participant annotated 8 unique sentences, with 2 annotations per condition, under 4 different conditions:

\begin{enumerate}
    \item \textbf{Manual Annotation with Excel:} Participants were provided with a spreadsheet containing source text, machine translation, and reference text. A color guide was used to annotate error categories and severities manually. More details of the instructions provided to participants can be found in Appendix~\ref{sec:appendix_excel}. 

    \item \textbf{\appname\ without Suggestions:} Participants used the \appname\ interface with no model-generated error detections.

    \item \textbf{\appname\ with \textit{XCOMET} Suggestions:} Participants received automatic error span suggestions from the XCOMET model, which were pre-highlighted in the interface.

    \item \textbf{\appname\ with \textit{EC-1} Suggestions:} Participants used the full system with GPT-4o-based error detection, which included both span highlighting and explanatory tooltips.
\end{enumerate}

\begin{figure}[h]
    \centering
    \includegraphics[width=1\linewidth]{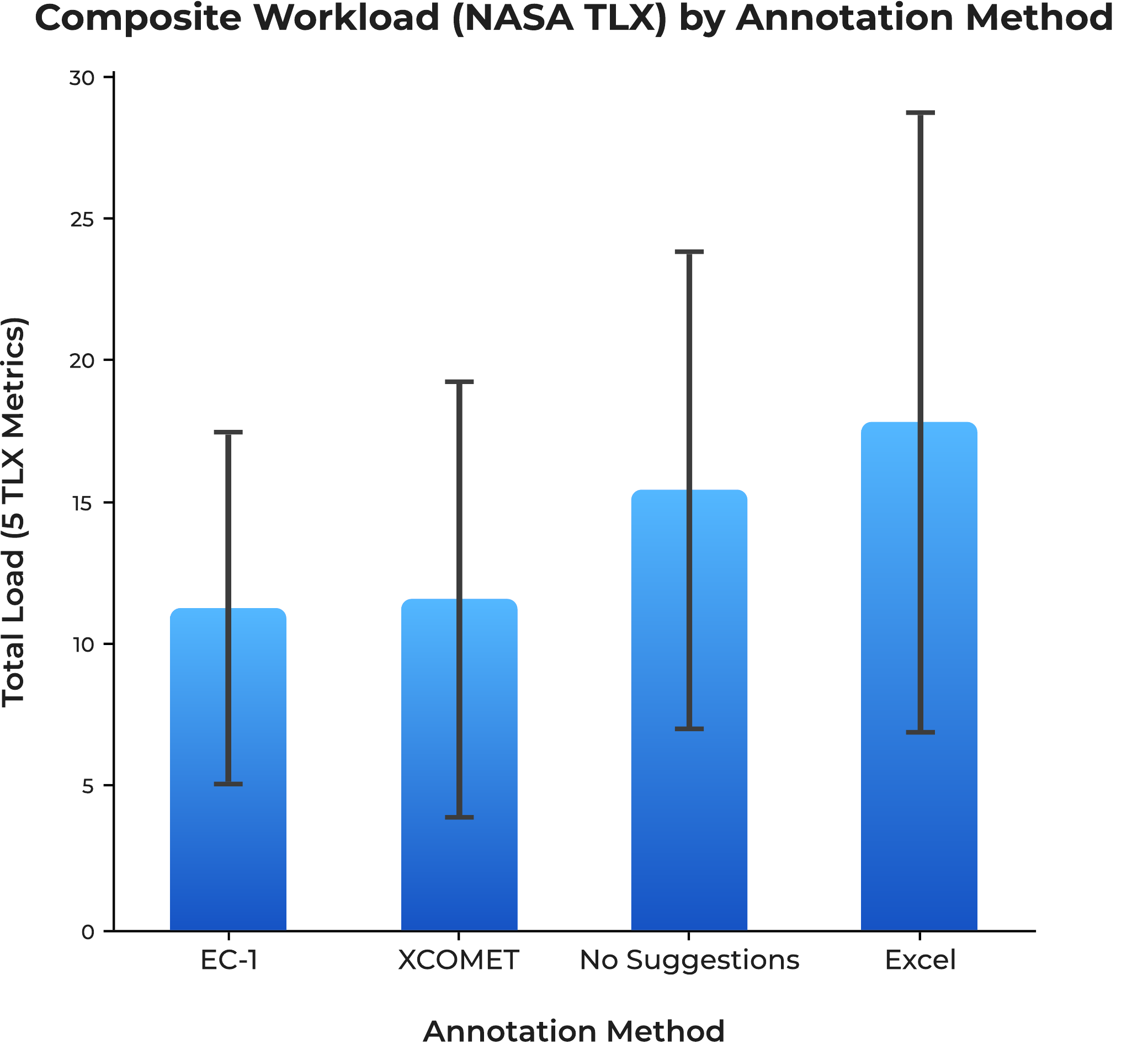}
    \caption{Composite Total Workload across Annotation Methods, calculated as the sum of five NASA TLX dimensions (Mental, Physical, Temporal, Effort, Frustration). Lower scores reflect reduced perceived workload.}

    \label{fig:Composite Workload}
\end{figure}

Figure~\ref{fig:Composite Workload} presents the composite workload scores across each annotation method in the study. The Excel manual annotation method shows the highest average workload, followed by the ``No Suggestions'' condition within \appname. Both error detection models, EC-1 and XCOMET conditions, demonstrated substantially lower average workload scores, indicating a reduction in cognitive burden on users. The error bars indicate considerable variability within workload ratings for the Excel and ``No Suggestion'' methods, while the EC-1 and XComet conditions exhibited more consistent results.

\noindent\textbf{Composite Workload Calculation.}  
The \emph{Total Load} in Figure~\ref{fig:Composite Workload} is computed as the simple sum of the five TLX dimensions—Mental Demand, Physical Demand, Temporal Demand, Effort, and Frustration—following NASA-TLX guidelines \cite{nasa-taskload-index}. We exclude Performance from this composite since it measures perceived success (higher is better), whereas the other five metrics indicate workload (lower is better). No additional weighting or post‐processing was applied.

Table~\ref{tab:nasa_tlx} presents the results of the user study from an HCI perspective. Across all NASA TLX dimensions, \appname consistently outperformed the manual Excel-based annotation method, demonstrating significant reductions in \textbf{mental demand}, \textbf{effort}, and \textbf{frustration}, while also improving \textbf{perceived performance}. These results demonstrate the effectiveness of our framework in streamlining translation workflows and alleviating the cognitive burden on annotators.

\begin{table*}[t!]
\centering
\arrayrulecolor{black}
\resizebox{0.8\textwidth}{!}{
\begin{tabular}{lrrrrrr}
\toprule
& Mental & Physical & Temporal & Effort & Frustration & Performance \\
\midrule
Mental      & 1.00 & 0.44 & 0.46 & 0.47 & 0.52 & -0.06 \\
Physical    & 0.44 & 1.00 & 0.49 & 0.34 & 0.30 & \cellcolor{red!20}-0.44 \\
Temporal    & 0.46 & 0.49 & 1.00 & 0.37 & 0.25 & -0.20 \\
Effort      & 0.47 & 0.34 & 0.37 & 1.00 & \cellcolor{blue!20}0.60 & -0.26 \\
Frustration & 0.52 & 0.30 & 0.25 & \cellcolor{blue!20}0.60 & 1.00 & -0.28 \\
Performance & -0.06 & \cellcolor{red!20}-0.44 & -0.20 & -0.26 & -0.28 & 1.00 \\
\bottomrule
\end{tabular}
}
\caption{Correlation Matrix of the NASA TLX Metrics}
\label{tab:nasa_tlx_correlation}
\end{table*}

To better understand the internal relationships between different workload factors, we computed Pearson correlation coefficients between TLX dimensions. As shown in Table~\ref{tab:nasa_tlx_correlation}, cognitive and emotional burdens—particularly \textbf{Mental Demand}, \textbf{Effort}, and \textbf{Frustration}—were positively correlated, confirming the internal consistency of the TLX framework in our study. \textbf{Perceived Performance} was negatively correlated with most workload dimensions, most notably with \textbf{Physical Demand} ($r = -0.44$), suggesting that reducing user effort and fatigue may directly contribute to greater perceived success. 

\subsubsection*{Statistical Analysis}
To assess significance across our four annotation methods \textit{(Excel, No Suggestions, XComet, EC-1)}, we applied the \textbf{Friedman test} to each NASA TLX dimension. Significant differences were found for \textbf{Mental Demand}, \textbf{Physical Demand}, and \textbf{Frustration} ($\chi^2(3) = 11.09$, \emph{p} = .011; $\chi^2(3) = 10.42$, \emph{p} = .015; $\chi^2(3) = 7.88$, \emph{p} = .049).

For focused comparisons between Excel and EC-1, we ran \textbf{Wilcoxon signed-rank tests}, which confirmed that EC-1 \textbf{significantly reduced}:
\begin{itemize}[noitemsep,nolistsep]
  \item \textbf{Mental Demand} ($W = 2.5$, \emph{p} = .010),
  \item \textbf{Physical Demand} ($W = 2.0$, \emph{p} = .041),
  \item \textbf{Frustration} ($W = 0.0$, \emph{p} = .027).
\end{itemize}

NASA TLX scores are ordinal and not normally distributed, making non-parametric tests appropriate. We thus used the Friedman test for within-subject comparisons across conditions, and Wilcoxon signed-rank tests for focused pairwise contrasts.

These results corroborate that our predictive‐error interface meaningfully lowers the annotator's cognitive and emotional workload compared to a standard spreadsheet baseline. These findings further support the HCI-driven design choices in \appname, such as predictive error suggestions and minimizing interface friction through quick action buttons corresponding to crucial post-editing tasks intended to reduce cognitive and physical load.

\section{Conclusions and Future Work}



In this work, we introduced \appname, a unified framework designed to streamline MT workflows while enhancing data collection for MT research. By integrating MT generation, error prediction, and translation post-editing within a single, user-friendly environment, \appname significantly improves translation efficiency and user satisfaction while annotating. Our framework also ensures that the annotated data collected from human annotators using our framework can be exported with state-of-the-art MT dataset standards, following MQM and ESA standards. As this paper focuses on annotation tooling, no accompanying dataset has been published. Conducting human annotations is a lengthy process, and we are working on creating a large and quality-assured dataset with \appname used for annotation. The benefits of our framework assist both translators by offering a seamless post-editing experience and researchers by providing high-quality, standardized datasets for fine-tuning current models, such as XCOMET and NLLB, as well as newer models that will be released in the future.

Empirical evaluation demonstrates that \appname outperforms traditional translation workflows, such as those annotation workflows based on Excel, in terms of both efficiency and user satisfaction. Our user study indicates that translators find our framework intuitive, efficient, and enjoyable, highlighting the importance of HCI considerations in our framework. 

\subsection{Continuous fine-tuning}





While our framework has already enhanced translation workflows, there is potential to incorporate continuous fine-tuning improvements into the underlying models when using our framework. One promising direction is to collect user-corrected data to fine-tune both the translation model (NLLB) and the error detection model (XCOMET). This additional feature would allow the system to dynamically improve based on the specific translation domain in which users are working, reducing the number of errors in the initial proposed translation and the number of errors detected by the error detection model.

Furthermore, as we have chosen NLLB and similar models as our translation model, alongside XCOMET as our error detection model, we can employ Low-Rank Adaptation \citep[LoRA]{hu2022lora} and other parameter-efficient strategies to carry out the fine-tuning process on limited compute. By integrating lightweight fine-tuning techniques, users could personalize their MT pipeline while maintaining efficiency on a local machine without needing to deploy anything on the cloud.

Nevertheless, the collection of data to carry out the continuous fine-tuning procedures remains difficult, thus, this direction remains a possible extension of our framework in the future.

\subsubsection*{Multimodal Extensions}
While our current framework is focused on text-based machine translation, we envision future extensions to support ASR (speech-to-text) and OCR (image-to-text) modalities. In such cases, the transcribed source (via ASR/OCR) would serve as input to the translation pipeline, followed by the same error detection and post-editing workflow. This would make the framework applicable to low-resource regions or archival content where text is not readily available. We leave implementation and evaluation of this multimodal pipeline for future work.

\section*{Limitations}



While our evaluation results demonstrate significant gains in translation efficiency and quality, some limitations remain:

\begin{itemize}[left=0mm]
\item Our user study was limited to 12 translators across 6 languages (Mandarin, Cantonese, Bengali, French, Japanese, Tamil), which may introduce sampling bias and limit generalizability. This evaluation was intended as a usability study to assess the effectiveness of the proposed framework, rather than a large-scale statistical evaluation.

    \item While \appname streamlines translation workflows, the final translation quality ultimately still depends on the skill and expertise of human annotators.


\item Although \appname supports low-resource MT models like NLLB, our current evaluation does not validate performance on low-resource languages. 

    \item Our custom GPT-4o assistant might not perform as expected when the source or target language is a low-resource language, as it is not trained intensively in those languages. 
    
\item The EC-1 assistant relies on OpenAI’s GPT-4o API, which may incur usage costs and raise data privacy concerns. Future work will explore open-weight LLMs such as Mixtral or LLaMA to mitigate these limitations.

\end{itemize}

By addressing these limitations, \appname has the potential to become an adaptive, user-driven translation framework, continuously improving through feedback while maintaining high usability and annotation efficiency. We hope that our framework will set a new standard for both translation workflows and MT data collection, bridging the gap between human expertise and machine translation systems.


\section*{Ethical Impact Statement}

The user study involved requesting the participants to carry out simple translation tasks and MT post-editing tasks, all in the \appname framework and Microsoft Excel. 

We have obtained approval from the Review Ethics Board to conduct the human study for our framework. The user study has active REB approval (File No: 18021).

\section*{Acknowledgments}

This research was supported by the Natural Sciences and Engineering Research Council of Canada (NSERC), Undergraduate Student Research Award (USRA). 



\nocite{rei-etal-2024-tower}
\nocite{semenov-etal-2023-findings-wmt-translation-terminologies}
\nocite{blain-etal-2023-findings-wmt-quality-estimation}
\nocite{huang-etal-2024-lost-llm-eval}

\bibliography{custom}

\clearpage

\appendix

\section{Related Works}
\label{sec:appendix_related_works}


\nocite{zhu-etal-2024-multilingual-llm-trans}
\nocite{wang-etal-2023-document-level-llm-trans}

NLLB \cite{nllbteam2022languageleftbehindscaling}, a translation model released by Meta AI, addresses translation task challenges by expanding translation capabilities to over 200 languages. NLLB demonstrates a significant advancement in MT performance over multiple metrics included in the WMT Shared Task \cite{freitag-etal-2023-wmt-results}. 

\nocite{silva-etal-2024-benchmarking-lrl-mt-nllb}

To evaluate the translation qualities of MT systems, metrics such as the MQM and ESA offer systematic approaches to analyze translation outputs. MQM \cite{burchardt-2013-multidimensional-mqm} is a comprehensive framework that categorizes translation errors based on predefined typologies and severity levels. MQM formalizes an analytic evaluation method by assigning translation errors to categories such as accuracy, fluency, and style, allowing for more thorough quality assessments. The framework has been widely adopted in the MT community with multiple variants \cite{blain-etal-2023-findings-wmt-quality-estimation, rei-etal-2020-comet, guerreiro-etal-2024-xcomet, kocmi-etal-2024-error-esa}, serving as one of the most widely used human evaluation metrics for MT tasks. 

While MQM offers detailed insights, it is time-consuming and often requires expert annotators, making large-scale evaluations and data collection costly and resource-intensive. To address these limitations, the ESA \cite{kocmi-etal-2024-error-esa} framework was introduced as a more efficient alternative. ESA combines elements of Direct Assessment \citep[DA]{bentivogli-etal-2018-machine-da} with error span marking alongside clear annotation instructions, enabling annotators to highlight specific error spans and assign severity scores. This method retains much of MQM's specificity while reducing the cognitive load on annotators, as it provides clear guidelines to distinguish between different errors, allowing for more efficient and meaningful data collection. Extensive studies by \citeauthor{kocmi-etal-2024-error-esa} show that ESA can match MQM’s effectiveness in system ranking while significantly reducing the time and expertise required for annotations.

As the demand for scalable MT evaluation grows, automatic metrics capable of providing interpretable and fine-grained assessments on MT outputs have gained more attention. XCOMET \cite{guerreiro-etal-2024-xcomet} represents a significant advancement in this domain by combining traditional sentence-level evaluation with detailed error span detection. Building on the foundations of earlier neural translation metrics like COMET \cite{rei-etal-2020-comet}, which focus on generating a single sentence-level quality score, XCOMET introduces the capability to detect and underline specific translation errors within a sentence. This improvement, specific to XCOMET, enables it to highlight error spans and assess their severity, in addition to a single sentence-level quality score, providing more interpretable evaluations that closely align with human evaluations. 


Recently, efforts were also placed into utilizing LLMs to provide a detailed analysis of translation errors. xTower \cite{treviso-etal-2024-xtower} is one such example that gives detailed descriptions and explanations of translation errors spans provided to the model. \citeauthor{treviso-etal-2024-xtower} have demonstrated that xTower can enhance the interpretability of translation errors identified by XCOMET. 

While tools like Appraise \cite{federmann-2018-appraise} and MT-EQuAl \cite{girardi-etal-2014-mt} remain widely used in shared tasks and research settings due to their lightweight interface and support for MQM-style annotations, they are limited in functionality. For example, Appraise does not offer predictive error suggestions or integrated LLM-based assistants. In contrast, our framework assists annotators through interactive error span detection and correction workflows powered by models such as XCOMET and GPT-4o, making it more suitable for real-time annotation and educational use.

Although Appraise has long supported structured MT annotation workflows, our study used Excel as a baseline because it reflects a widely used but friction-heavy process many annotators and researchers may adopt due to a lack of access to specialized annotation tools. We selected Excel to represent a realistic baseline for comparison. Future work may evaluate our framework more directly against Appraise and other task-specific tools to assess annotation quality and efficiency in more detail.






\section{Standardized Error Definition and Ruleset}
\label{sec:appendix_error_definition}

In our study, we define several error categories to assess the quality of translations. \appname allows the annotator to categorize any error spans under these given categories, enabling them to easily select one category to associate with an error span. 

To avoid having too many categories with similar definitions, and to ensure that each error category is distinct and easily identifiable given an error span, we have simplified the existing categories that MQM \cite{burchardt-2013-multidimensional-mqm} provides into the following:

\begin{itemize}[left=0mm]
    \itemsep 0em

    \item Addition, where content that is not present in the target text appears in the source. 

    \item Omission, which refers to content from the source that is missing in the target

    \item Mistranslation, where the target text inaccurately represents the source content

    \item Untranslated, where a segment intended for translation is omitted

    \item Grammar, which covers violations of grammatical rules in the target language

    \item Spelling, where words are misspelled
    
    \item Typography, which addresses visual presentation issues such as incorrect punctuation, inconsistent capitalization, or spacing errors
    
    \item Unintelligible, where the text is garbled or incomprehensible
\end{itemize}

\section{Custom GPT-4o assistant dubbed EC-1}
\label{sec:appendix_gpt_4o}

To supplement traditional error detection models like XCOMET, we implemented a custom GPT-4o assistant named \textbf{EC-1}. This assistant is designed to analyze translation outputs with detailed reasoning and character-level span annotations, offering a more interpretable alternative for translation error detection and annotation.

\paragraph{Prompt Design}
EC-1 is prompted as a professional linguist specializing in machine translation evaluation. Given a source sentence and its corresponding machine translation, EC-1 is instructed to:
\begin{itemize}
    \itemsep 0em
    \item Detect fine-grained translation errors.
    \item Label each error with a type from a predefined taxonomy: \textit{Addition, Omission, Mistranslation, Untranslated, Grammar, Spelling, Typography, Unintelligible}.
    \item Assign each error a severity level: \textit{Minor} or \textit{Major}.
    \item Provide precise, non-overlapping character-level spans in both source and translation texts.
    \item Justify each detected error with a brief explanation.
\end{itemize}

The assistant returns structured, ESA-compatible JSON output for each error. This format is directly compatible with our annotation interface and error span alignment.

\paragraph{Example Use}
A sample input passed to the EC-1 API is structured as follows:
\begin{CJK}{UTF8}{min}
\begin{quote}\small
\texttt{Source: "Today Romani is spoken by small groups in 42 European countries."}\\
\texttt{MT: "Todayen Romani は欧州の42か国で小グループで語られています."}
\end{quote}
\end{CJK}

EC-1 returns:

\begin{quote}\footnotesize
\begin{verbatim}
{
  "error_spans": [
    {
      "original_text": "Today",
      "error_type": "Spelling",
      "error_severity": "Minor",
      "start_index_orig": 0,
      "end_index_orig": 5,
      "start_index_translation": 0,
      "end_index_translation": 7,
      "correct_text": "The word 'Today' is
      incorrectly rendered as 'Todayen'..."
    },
    ...
  ]
}
\end{verbatim}
\end{quote}

Our prompt emphasizes:
\begin{itemize}
    \item Non-overlapping spans,
    \item Strict 0-based character indexing,
    \item A consistent structure aligned with MQM and ESA principles.
\end{itemize}

EC-1 responses are integrated directly into the \appname interface, offering users interpretable, guided suggestions for post-editing.

\section{Excel Annotation Instructions}
\label{sec:appendix_excel}

To assess the efficacy of traditional annotation methods, we have designed a ruleset for the user study participants to annotate on the given test entries. 

\begin{figure}
    \centering
    \includegraphics[width=0.5\linewidth]{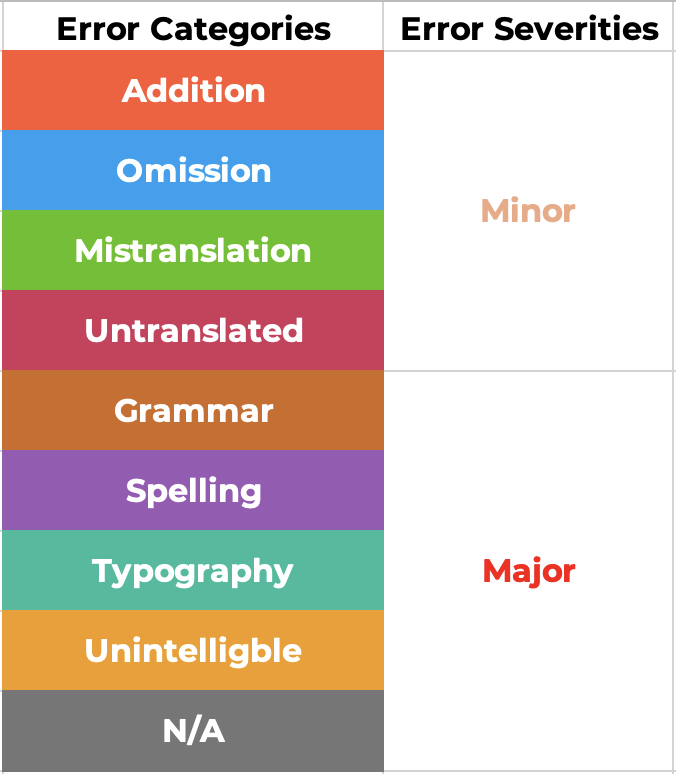}
    \caption{Format that was given to annotators to annotate our test entries with}
    \label{fig:excel-format}
\end{figure}

As shown by Figure~\ref{fig:excel-format}, we have asked the annotators to highlight text using multiple different colors to indicate different error categories, as outlined in Appendix~\ref{sec:appendix_error_definition}. The annotators are also told to use bold font to indicate if the identified error is a Major error, and a non-bold font to indicate that the error is a minor error. 

\section{User Study Details}
\label{sec:appendix_user_study_details}

We collected our user study data through Google Forms, created the survey using a standard NASA TLX format, and exported user submissions to a CSV format. We then performed statistical analyses on the collected data programmatically using Python and its scientific and numerical packages. A sample of the form \footnote{{https://forms.gle/NJcNSPyEBfSUKMVC8}} used to collect the data in the user study is available. Participants were volunteer student annotators who were native speakers of the respective non-English language they were annotating and were not involved in the authorship of this paper. No monetary compensation was provided.



\end{document}